\pdfoutput=1
\documentclass[11pt]{article}
\usepackage{ACL2023}  
\usepackage{lineno}  
\usepackage{times}
\usepackage{latexsym}
\usepackage[T1]{fontenc}
\usepackage[utf8]{inputenc}
\usepackage{microtype}
\usepackage{inconsolata}
\usepackage{graphicx}
\usepackage{multirow}
\usepackage{verbatim}

\usepackage{float}
\makeatletter

\renewcommand{\linenumbers}{%
  \ifodd\thepage \rightlinenumbers \else \leftlinenumbers \fi}
\makeatother

\linenumbers

\title{Refining Syntactic Distinctions Using Decision Trees: A Paper on Postnominal 'That' in Complement vs. Relative Clauses}

\author{Hamady GACKOU \\
  Student at Paris Cit\'e University, 1 Rue Pajol, 75018 Paris, France \\
  \texttt{hamady.gackou@etu.u-paris.fr}}

\begin{document}
\maketitle

\begin{abstract}
In this study, we first tested the performance of the TreeTagger English model developed by Helmut Schmid with test files at our disposal, using this model to analyze relative clauses and noun complement clauses in English. We distinguished between the two uses of "that," both as a relative pronoun and as a complementizer. To achieve this, we employed an algorithm to reannotate a corpus that had originally been parsed using the Universal Dependency framework with the EWT Treebank. In the next phase, we proposed an improved model by retraining TreeTagger and compared the newly trained model with Schmid's baseline model. This process allowed us to fine-tune the model's performance to more accurately capture the subtle distinctions in the use of "that" as a complementizer and as a nominal. We also examined the impact of varying the training dataset size on TreeTagger's accuracy and assessed the representativeness of the EWT Treebank files for the structures under investigation. Additionally, we analyzed some of the linguistic and structural factors influencing the ability to effectively learn this distinction.
\end{abstract}

\section{Introduction}
\subsection{Aim of the Study}
This study aims to explore the syntactic distinctions in the usage of the postnominal "that" in English subordinate clauses, focusing specifically on complement and relative clauses. Our primary objective is to refine the understanding of how "that" functions in these two distinct syntactic contexts—first as a relative pronoun in relative clauses, and second as a complementizer in noun complement clauses.

\subsection{Context and Motivation}
Understanding the syntactic behavior of "that" is crucial for computational linguistics and natural language processing (NLP) as it enables more accurate syntactic parsing and semantic interpretation. While previous research has investigated the syntactic roles of various linguistic markers, the distinction between complementizer and relative pronoun use of "that" remains a nuanced challenge \cite{Ballier2003-Title1, Ballier2010-Title2, Schmid2023-Title3, Author2022-Title4, Author2018-Title5}. Current syntactic models, including those built on frameworks such as Universal Dependencies and Treebank annotations, struggle with accurately distinguishing between these two roles in real-world data. This gap in linguistic modeling is addressed by applying an enhanced algorithm to reannotate corpora and improve parsing accuracy.

Moreover, research on NLP models, such as the TreeTagger English model developed by Schmid \cite{Schmid2023-Title3}, highlights the limitations of existing tools when applied to complex syntactic structures. The motivation behind this study lies in optimizing these tools to achieve a finer granularity of analysis for syntactic distinctions, particularly in ambiguous contexts involving postnominal "that."

\subsection{Research Questions}

Understanding the syntactic distinction between the relative pronoun and complementizer uses of \textit{that} in English subordinate clauses remains a challenge for automated parsing systems. This study investigates:

\begin{itemize}
    \item How can we reliably distinguish between the relative pronoun and complementizer uses of \textit{that} in English subordinate clauses?
    \item What impact does retraining the TreeTagger model have on its ability to differentiate these two syntactic roles?
    \item How does the size and quality of the training dataset influence the accuracy of syntactic parsing for these structures?
    \item Is the EWT Treebank a representative dataset for capturing the subtleties in the use of \textit{that}?
\end{itemize}

Linguistic research has shown that dependency patterns in complex sentences can provide insights into syntactic role assignment \cite{Wang2019}. Previous studies have examined the ambiguity of postnominal \textit{that} and proposed computational models for its disambiguation \cite{Ballier2010-Title2, tighidet2022}. Additionally, the use of decision trees in probabilistic part-of-speech tagging has demonstrated improvements in syntactic analysis \cite{Schmid1994, Schmid1995}. Retraining taggers like TreeTagger \cite{Schmid2023-Title3} with annotated corpora, such as those used in dependency parsing studies \cite{Zeldes2017}, may enhance disambiguation performance. However, the representativeness of corpora like the EWT Treebank for such distinctions remains an open question \cite{Ballier2003-Title1, Author2018-Title5}.

\subsection{Theoretical Background}

In English, the word \textit{that} plays a crucial role in forming subordinate clauses, particularly as a relative pronoun introducing relative clauses and as a complementizer in nominal complement clauses. For example, in the sentence \textit{"The book that I read was fascinating"}, \textit{that} is a relative pronoun introducing a relative clause modifying \textit{book}. Conversely, in \textit{"She believes that he is honest"}, \textit{that} functions as a complementizer introducing a nominal complement clause. Although both structures involve \textit{that}, they differ in their syntactic and semantic properties. Relative clauses generally modify a noun and provide additional information, whereas nominal complement clauses act as complements to a verb, noun, or adjective, expressing the content of a belief, statement, or perception.

\subsection{Problem of Reannotation}
 Distinguishing \textit{that} as a relative pronoun from \textit{that} as a complementizer presents significant challenges in syntactic annotation, particularly in corpora following the Universal Dependencies (UD) framework \cite{Zeldes2017, Author2018-Title5}. In UD, relative clauses are typically annotated with the relation \texttt{acl:relcl} (adnominal clause: relative clause), while nominal complement clauses are annotated with relations such as \texttt{ccomp} (clausal complement). However, the distinction between these two uses of \textit{that} is not always clear, especially in complex or ambiguous sentences \cite{tighidet2022, Ballier2010-Title2}.

Several studies have highlighted the difficulties of maintaining annotation consistency in large-scale corpora, particularly for postnominal \textit{that} \cite{Author2018-Title5, gaillat2016}. This ambiguity poses challenges not only for human annotators but also for dependency parsers, leading to inconsistencies that affect downstream NLP applications such as syntactic parsing, machine translation, and information retrieval \cite{Wang2019}. While probabilistic models and decision-tree-based part-of-speech tagging have shown promise in refining these distinctions \cite{Schmid1994, Schmid1995}, annotation quality remains a crucial bottleneck.

To address these issues, this study investigates how retraining a part-of-speech tagger like TreeTagger on a specifically annotated corpus can improve the distinction between these two syntactic roles. By refining syntactic annotation and leveraging decision-tree-based approaches, we aim to enhance the robustness of parsing systems when dealing with postnominal \textit{that}.

\subsection{Previous Work on "That" Analysis and Reannotation}

Several studies have examined the distinction between \textit{that} as a relative pronoun and as a complementizer. For example, Tighidet and Ballier (2022) explored two methods to analyze relative and nominal complement clauses in English, using distinct labels for \textit{that} based on its function. They applied an algorithm to relabel a corpus annotated with the GUM Treebank under Universal Dependencies and used TreeTagger to train a model to distinguish between the two uses of \textit{that}. Their study demonstrated that reannotation and model training with clearer function-based distinctions could enhance syntactic parsing accuracy \cite{tighidet2022}.

Furthermore, Gaillat (2016) investigated constructions involving \textit{this} and \textit{that} among French- and Spanish-speaking learners of English, emphasizing the need for more refined reannotation using functional labels and semantic positions. This research highlighted the importance of functional distinction in analyzing deictic and proform constructions and proposed a multi-level annotation framework for different corpora \cite{gaillat2016}.

\subsection{Tagging Schemes: CLAWS8 and Penn Treebank}

The CLAWS8 tagging system and the Penn Treebank are widely used resources for syntactic annotation in English. CLAWS8 is a grammatical tagging system that assigns detailed morphosyntactic labels to words, facilitating the analysis of complex syntactic structures. The Penn Treebank, on the other hand, is a syntactically annotated corpus that has served as a benchmark for numerous studies in computational linguistics. However, these resources may present limitations in distinguishing the various uses of \textit{that}. For instance, Gaillat (2016) emphasized the need for finer reannotation in these corpora to capture functional distinctions between \textit{that} as a relative pronoun and as a complementizer \cite{gaillat2016}.

\subsection{Universal Dependencies for Dependency Parsing}
The Universal Dependencies (UD) framework provides a consistent annotation scheme for syntactic relations across languages. In this framework, relative clauses are annotated with the \texttt{acl:relcl} (adnominal clause: relative clause) relation, while complement clauses are annotated with relations such as \texttt{ccomp} (clausal complement). The distinction between these relations is crucial for accurate syntactic parsing. For example, in the sentence \textit{"The book that I read was fascinating"}, \textit{that} introduces a relative clause modifying \textit{book} and would be annotated with \texttt{acl:relcl}. In contrast, in \textit{"She believes that he is honest"}, \textit{that} introduces a complement clause, annotated with \texttt{ccomp}.

\section{Materials and Methods}

The TreeTagger model was evaluated using two different Treebank models: the Penn Treebank (\texttt{penn.par}) and the BNC Treebank (\texttt{bnc.par}). These models were  evaluated on test corpora consisting of 200 sentences each. The focus of the evaluation was on the performance of the word \textit{that} across various categories, using the CLAWS5 and Penn Treebank tagsets.

\subsection*{2.1 Penn Treebank Model (\texttt{penn.par})}
The Penn Treebank model uses parameters from the Schmid Penn Model and is based on the Penn Treebank tagset. The evaluation was carried out for various categories such as CC (Coordinating Conjunctions), DT (Determiners), IN (Prepositions and Conjunctions), RB (Adverbs), WDT (Wh-determiners), and WP (Wh-pronouns). The precision, recall, and F1-score were computed for each of these categories using the TreeTagger tool.

\subsection*{2.2 BNC Treebank Model (\texttt{bnc.par})}
The BNC Treebank model was used with the CLAWS5 tagset, which is widely used in British English corpora. For the word \textit{that}, categories such as \texttt{n\_CJT} (Conjunctions), \texttt{n\_DTO}, \texttt{n\_DTQ}, \texttt{n\_NULL}, \texttt{n\_PNQ}, and \texttt{n\_VBZ} were analyzed. The performance was measured in terms of precision, recall, and F1-score.

\subsection*{2.3 Preprocessing the Corpus for CLAWS C8 Tagset Compatibility}

In this section, I outline the detailed steps I followed to adapt the Brown Corpus to the CLAWS C8 tagset. This process involved several essential stages, which I will describe in detail.

\subsection*{2.4 Downloading and Extracting the Brown Corpus}
The Brown Corpus, consisting of 500 small files, was downloaded using the built-in NLTK module. Each file was saved as a .txt file in a dedicated directory rawbrowndata, allowing for easy access and processing. The code used to download and save the corpus is shown below.

{\scriptsize
\begin{verbatim}
import nltk
from nltk.corpus import brown
import os

nltk.download('brown')
sentences = brown.sents()
output_folder = "raw_brown_data"
if not os.path.exists(output_folder):
    os.makedirs(output_folder)

for i, file_id in enumerate(brown.fileids()):
    file_text = " ".join(brown.words(file_id))
    output_file = os.path.join(output_folder, f"{file_id}.txt")
    with open(output_file, "w", encoding="utf-8") as f:
        f.write(file_text)
    print(f"Saved {file_id} -> {output_file}")
\end{verbatim}

}

\subsection*{2.5 Annotation with UDPipe}
Once the raw corpus was prepared, I proceeded with annotating each file using the UDPipe REST API. This step is crucial to transforming the raw text into a structured format (CoNLL-U), which includes tokenization, part-of-speech (POS) tagging, and dependency parsing. The annotation was performed using the \textit{english-ewt} model provided by UDPipe.

The code used for this annotation is as follows:
{\scriptsize 
\begin{verbatim}
import requests
def annotate_text_with_udpipe(text, model="english-ewt"):
    url = "https://lindat.mff.cuni.cz/services/udpipe/api/process"
    payload = {
        "data": text,
        "model": model,
        "tokenizer": "yes",
        "tagger": "yes",
        "parser": "yes"
    }
    response = requests.post(url, data=payload)
    if response.status_code == 200:
        return response.json()["result"]
    else:
        raise Exception(f"UDPipe API Error: {response.status_code}, {response.text}")
\end{verbatim}
} %

\subsection*{2.6 Applying Heuristic for CLAWS C8 Tagset Adaptation}
Once the corpus was annotated, the next step was to apply a heuristic method to adjust the annotations according to the CLAWS C8 tagset, focusing particularly on reannotating the occurrences of the word "that." Depending on the context, "that" was assigned either the WPR tag (relative pronoun) or the CST tag (subordinating conjunction). This heuristic involved analyzing the dependency relations and POS tags for each occurrence of "that."

The code for applying the heuristic is shown below:

{\scriptsize 
\begin{verbatim}
def deps_emulator(folder_name="brown_annotated/"):
    sentences_reannotated_WPR = []
    sentences_reannotated_CST = []
    # Other processing steps...
    if sentences[i][j]['form'].lower() == 'that':
        # Reannotation based on context
        # WPR (relative pronoun) or CST (subordinating conjunction)
\end{verbatim}
} %

\subsection*{2.7 Displaying Annotated Sentences and Verifying Results}
After the heuristic reannotation, the sentences were examined. The \texttt{pprint} function was used to display the sentences, highlighting where "that" had been reannotated. This allowed me to manually check the annotations and ensure their accuracy.

Here is the code used to display the results:

{\scriptsize 
\begin{verbatim}
def correct_helper():
    print("WPR", "-"*100)
    n = 0
    for sent in sentences_WPR:
        print(n, pprint(sent))
        print()
        n += 1
    print("CST", "-"*100)
    n = 0
    for sent in sentences_CST:
        print(n, pprint(sent))
        print()
        n += 1
\end{verbatim}
} %

\subsection*{2.8 Final Steps and Preparing the Results}
After reannotating and verifying the results, the final corpus was structured into a format ready for training my model. This included saving the results into separate files and preparing the corpus for TreeTagger training.

Here is the code to finalize the corpus preparation:
 {\scriptsize 
\begin{verbatim}
 sentences_WPR, sentences_CST = deps_emulator().values()
\end{verbatim}
} %

\subsection*{Remark}
By the end of these steps, I successfully prepared an annotated corpus adapted to the CLAWS C8 tagset, which will be used for training my tagger model. This rigorous process, from obtaining the raw corpus to applying heuristics and verifying the results, ensures that the annotations are of high quality and ready for use in training machine learning models. This work provides a solid foundation for my future research and for training natural language processing models.

\subsection*{2.10 Revisiting the EWT Treebank for 'that' Annotation per Categories}

In this study, the distribution of the word \textit{'that'} in the Brown Corpus, specifically in the EWT Treebank, was analyzed and re-annotated for better syntactic clarity. The total number of occurrences of the word \textit{'that'} in all files was 10,788. After re-annotation, 974 instances of \textit{'that'} were revised, focusing on its categorization into two primary grammatical roles: **CST** (subordinating conjunction) and **WPR** (relative pronoun). 

The re-annotation revealed the following distribution:
- \textbf{551 instances} of \textit{'that'} were re-annotated as CST (subordinating conjunction).
- \textbf{423 instances} of \textit{'that'} were re-annotated as WPR (relative pronoun).

These findings emphasize the important syntactic distinction between \textit{'that'} acting as a subordinating conjunction and as a relative pronoun, reflecting the diverse syntactic environments in which \textit{'that'} occurs. Furthermore, the proportion of these occurrences in relation to the total number of \textit{'that'} instances is notable.

\subsection*{Verbal Relations}

The analysis also examined the occurrences of \textit{'that'} in specific dependency relations such as **acl** (adjectival clause) and **acl:relcl** (relative clause). Specifically, the number of verbs classified as **acl:relcl** without the presence of \textit{'that'} was counted as \textbf{955}. These represent verbs that are involved in relative clause constructions without the direct use of \textit{'that'}. In terms of proportional representation:
- The proportion of \textit{acl} verbs without \textit{'that'} is \textbf{0.0088\%}.
- The proportion of \textit{acl:relcl} verbs is slightly higher, at \textbf{0.0091\%}.

These findings provide an interesting perspective on how the presence of \textit{'that'} influences the syntactic structure of relative clauses and adjectival clauses.

\subsection*{Frequency of RC and NCC in the EWT Treebank}

Relative Clauses (RC) and Noun Complement Clauses (NCC) are crucial structures in the syntax of \textit{'that'}. In the EWT Treebank, \textit{'that'} is often involved in these constructions. This section explores the frequency of \textit{'that'} in both types of clauses and presents an analysis of their distribution across the dataset.

\begin{figure}[htbp]
    \centering
    \includegraphics[width=0.2\textwidth]{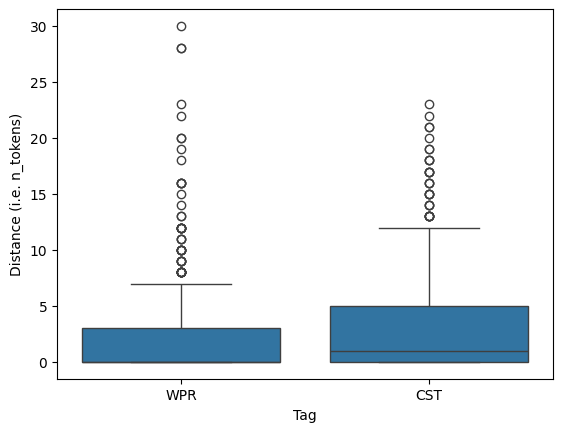}
    \caption{Illustration of the distribution of 'that' across categories in the Brown Corpus. This figure shows the proportions of \textit{'that'} instances re-annotated as CST and WPR, as well as their occurrence in different syntactic roles.}
    \label{fig:acl-rc}
\end{figure}
\subsection*{2.11 Lexicon Creation from Annotated Files}

A lexicon is a collection of words or terms used in a specific language or context, often with their meanings, grammatical categories, and other relevant linguistic information. It serves as an essential resource for natural language processing (NLP) and linguistic analysis, enabling machines to understand and generate human language. Lexicons are crucial for tasks such as part-of-speech tagging, word sense disambiguation, and syntactic parsing \cite{Manning1999}.

The creation of a lexicon typically involves extracting linguistic data from annotated corpora, allowing for a structured representation of terms and their associated features. In this study, we focus on the construction of lexicons based on re-annotated data from the Brown Corpus, specifically targeting the categorization of words such as *that* into their syntactic and semantic roles. The following procedure outlines the steps taken to create these lexicons.

\subsection*{Methodology for Lexicon Creation}

The process begins with the collection of re-annotated files from the Brown Corpus, focusing on specific syntactic categories. To group and prepare the data for lexicon construction, a script is used to concatenate the first *n* annotated files into a single lexicon file. The following pseudo code function illustrates this process:
{\scriptsize
\begin{verbatim}
# Pseudocode for concatenating n files into a single output file:

# Define function to concatenate first n files
function concat_n_first(n):
    c = 0
    
    # Create output directory if it doesn't exist
    output_dir = "brown_annotated/TOKEN_PER_ROW/GROUPED_FILES"
    if not directory_exists(output_dir):
        create_directory(output_dir)
    
    # Open the output file
    open_output_file(f"{output_dir}/{n}.txt")
    
    # Iterate over all files in the source directory
    for each file in directory("brown_annotated/TOKEN_PER_ROW"):
        if file_is_valid(file):
            open_input_file(file)
            content = read_file_content(file)
            write_to_output(content)
            c += 1
            
        # Stop after processing n files
        if c == n:
            break
\end{verbatim}
}

This script consolidates the content of the first *n* files into a single file, which is then used to populate the lexicon with the relevant linguistic information. The resulting lexicon can be utilized for further analysis and processing in downstream NLP tasks.

\begin{figure}[htbp]
    \centering
    \includegraphics[width=0.3\textwidth]{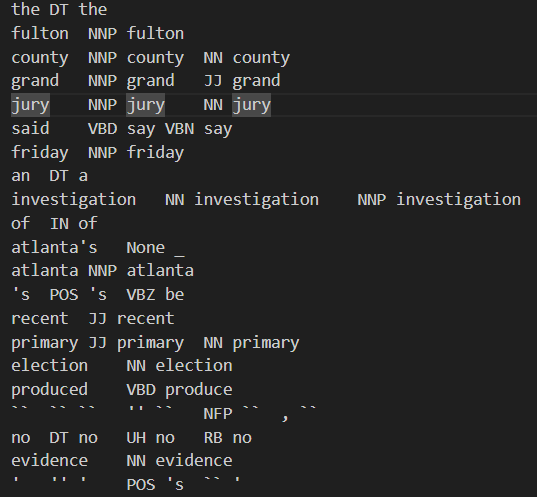}
    \caption{An example of the structure of the lexicon created from the Brown Corpus.}
    \label{fig:lexicon_example}
\end{figure}

\subsection*{2.12 Re-Training with TreeTagger}

TreeTagger is a widely used tool for part-of-speech tagging and other types of annotation in natural language processing (NLP). Re-training a model with TreeTagger involves providing annotated training data to create a model capable of classifying tokens in a specific corpus according to syntactic categories. This process is essential for fine-tuning the model to handle specialized data or domains, such as annotated corpora for a particular language or syntactic structure \cite{Schmid1995}.

The re-training procedure involves several stages: preparing annotated files in a structured format, grouping them into lexicons, and feeding them into the TreeTagger training process. The following commands illustrate the re-training process for different sizes of lexicons:
{\scriptsize
\begin{verbatim}
!train-tree-tagger -st . brown_annotated/TOKEN_PER_ROW/GROUPED_FILES/lexicon_10.txt 
empty.txt brown_annotated/TOKEN_PER_ROW/GROUPED_FILES/10.txt brown_annotated/TOKEN_PER_ROW/GROUPED_FILES/pen_cw8_model_10.par
\end{verbatim}
}
\subsection*{ Process explanation}
The command `train-tree-tagger` trains the model using different lexicon sizes, indicated by the number of tokens (e.g., 10, 30, 100, 200, 300, 500). The result is a trained model stored as a .par file (e.g., `pencw8model10.par`), which can be used for tagging new data.
The process of re-training TreeTagger is critical for improving the accuracy of syntactic analyses, as it enables the model to adapt to the specific characteristics of the training data. By gradually increasing the size of the training set, the model can capture a broader range of syntactic patterns, leading to improved performance on downstream NLP tasks \cite{Schmid2004, Nivre2006}.

\begin{figure}[htbp]
    \centering
    \includegraphics[width=0.3\textwidth]{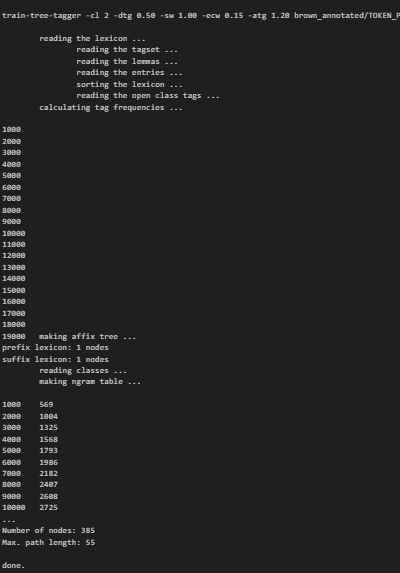}
    \caption{Illustration of the TreeTagger training process}
    \label{fig:tree_tagger_training}
\end{figure}

\subsection*{Theoretical Background}

Re-training a part-of-speech tagger such as TreeTagger relies on supervised learning, where a model is trained using manually annotated data. This process is central to improving tagger accuracy, especially in cases where domain-specific linguistic phenomena need to be captured \cite{Goldberg2017, Kudo2005}.

The effectiveness of re-training depends on several factors, including the quality and quantity of the annotated training data, the specific features used in training, and the algorithm employed. TreeTagger uses a decision tree algorithm that classifies words based on their context and syntactic features \cite{Schmid1995}. The larger the lexicon and the more diverse the training data, the better the model can generalize to unseen data.

\subsection*{2.13 Test File Formation for CST and WPR Categories}

In this section, we describe the procedure used to create test files for the categories CST and WPR, focusing on the usage of the word "that" in different syntactic contexts.

\subsection*{Creating Test Files for WPR Category}
For the WPR category, "that" is used as a relative pronoun, often introducing a relative clause. To generate relevant data, we constructed a set of 200 sentences where "that" serves as a relative pronoun. These sentences are designed to capture the syntactic role of "that" in defining and non-defining relative clauses. The goal was to collect a sufficient sample that would allow for robust testing of the model's ability to correctly identify and label "that" in the WPR context.

\textit{Prompt Example for WPR:}  
\begin{quote}
"Give me two hundred sentences where 'that' is used as a relative pronoun, introducing a relative clause."
\end{quote}

\subsection*{Creating Test Files for CST Category}
For the CST category, "that" functions as a complementizer, serving as a conjunction to introduce a complement clause. Similarly to the WPR category, we created a set of 200 sentences in which "that" is used as a complementizer. These sentences are structured to include "that" in contexts where it introduces noun clauses, such as reported speech or clauses serving as the object of a verb. This sample ensures that the model can effectively recognize and classify "that" as a complementizer in different syntactic configurations.

\textit{Prompt Example for CST:}  
\begin{quote}
"Give me two hundred sentences where 'that' is used as a complementizer, introducing a complement clause."
\end{quote}

Both test file sets, for WPR and CST, are essential for evaluating the model's performance in accurately distinguishing between these two syntactic roles of "that" and for refining the model's tagging accuracy in context-sensitive scenarios.


\section{3. Results}

\subsection*{3.1. Analysis of UD EWT Treebank}

An exploration of the EWT Treebank annotated under the UD scheme revealed several limitations in distinguishing between relative clauses and nominal complement clauses introduced by \textit{that}. Some annotations fail to clearly differentiate the two structures, leading to syntactic ambiguities. For instance, cases were observed where \textit{that} is annotated identically regardless of its specific function within the sentence. Moreover, the representation of dependency relations, such as \texttt{acl:relcl} for relative clauses and \texttt{ccomp} for complement clauses, is not always consistent, which can affect NLP models trained on these data. Extracting specific data on the occurrence and distribution of these structures highlighted the need for reannotation to improve accuracy and consistency, particularly for complex sentences where the function of \textit{that} may be ambiguous.
The analysis of the EWT Treebank annotated under the UD scheme revealed limitations in distinguishing relative clauses and nominal complement clauses introduced by \textit{that}, leading to syntactic ambiguities. Notably, 99\% of acl instances occur in a left-to-right order with an average distance of 2.9 between parent and child. The acl relation predominantly connects NOUN-VERB pairs (88\%), with smaller percentages involving other part-of-speech combinations. These inconsistencies highlight the need for reannotation to improve the accuracy of syntactic distinctions, especially for complex sentences involving \textit{that}.

For a visual representation, see the following  :

\begin{figure}[htbp]
    \centering
    \includegraphics[width=0.2\textwidth]{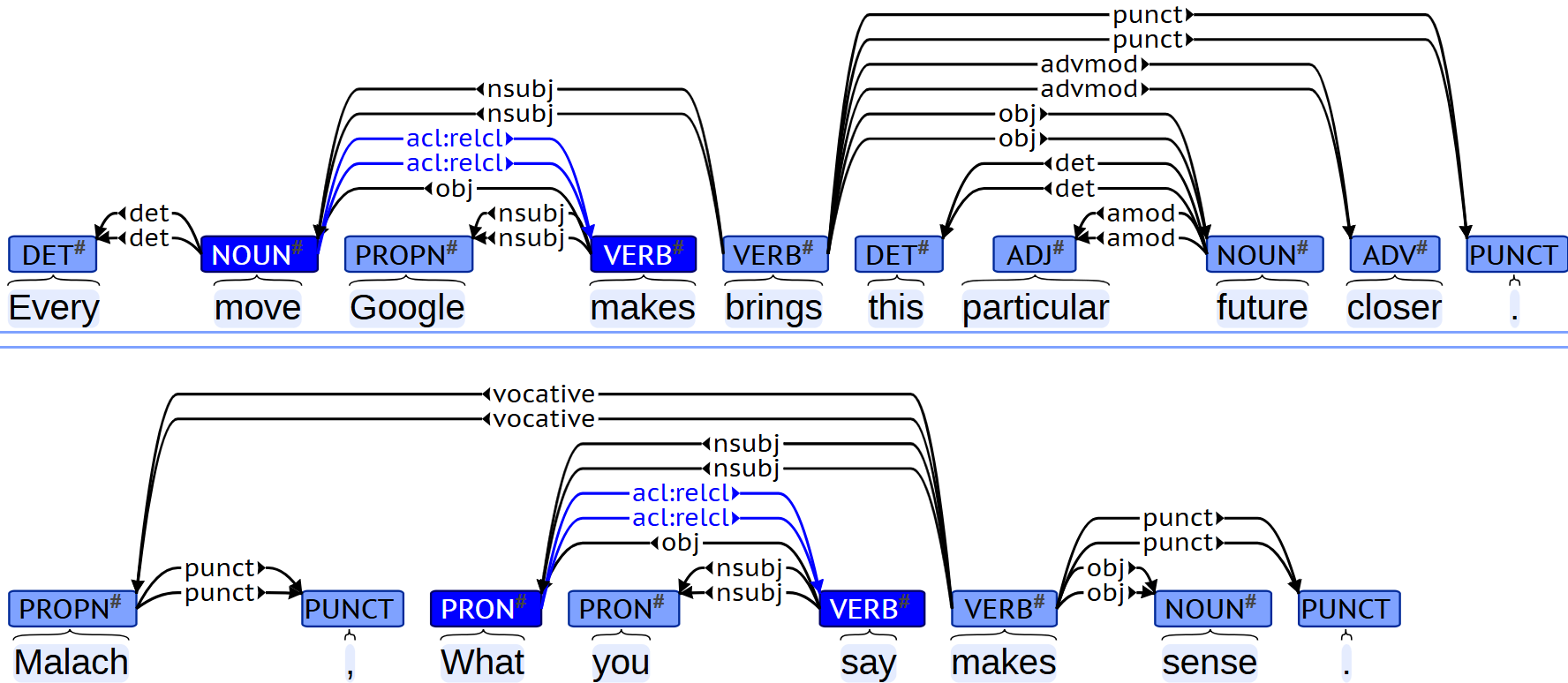}
    \caption{Illustration of representation .}
    \label{fig:acl-ewt}
\end{figure}

\subsection*{3.2. Penn Treebank Model (\texttt{penn.par}) Results}
The performance of TreeTagger with both the Penn Treebank and BNC Treebank models was evaluated in terms of precision, recall, and F1-score for the word \textit{that}. The results for each tagset and category are summarized below.
The performance results of the Penn Treebank model for the word \textit{that} are as follows:

\begin{itemize}
    \item \textbf{CC (Coordinating Conjunctions)}: The model did not recognize any instances of the category, leading to null results for all three metrics (precision, recall, F1-score).
    \item \textbf{DT (Determiners)}: Precision = 0.6, Recall = 0.6, F1-score = 0.6. Moderate performance was achieved in identifying determiners.
    \item \textbf{IN (Prepositions and Conjunctions)}: Precision = 0.36, Recall = 0.36, F1-score = 0.36. The model struggled to identify this category accurately.
    \item \textbf{RB (Adverbs)}: Precision = 0.0, Recall = 0.0, F1-score = 0.0. The model did not perform well for adverbs.
    \item \textbf{WDT (Wh-determiners)}: Precision = 0.04, Recall = 0.04, F1-score = 0.04. The model struggled to recognize wh-determiners.
    \item \textbf{WP (Wh-pronouns)}: Precision = 0.0, Recall = 0.0, F1-score = 0.0. The model failed to recognize wh-pronouns.
\end{itemize}

\subsection*{3.3. BNC Treebank Model (\texttt{bnc.par}) Results}
The performance of TreeTagger with the BNC Treebank model for the word \textit{that} is as follows:

\begin{itemize}
    \item \textbf{n\_CJT (Conjunctions)}: Precision = 1.0, Recall = 1.0, F1-score = 1.0. The model performed flawlessly in identifying conjunctions.
    \item \textbf{Other categories (n\_DTO, n\_DTQ, n\_NULL, n\_PNQ, n\_VBZ)}: No predictions were made for these categories, leading to zero precision, recall, and F1-scores.
\end{itemize}

\subsection*{4. Comparative Analysis of Models}
A comparison between the Penn Treebank and BNC Treebank models reveals notable differences:

\begin{itemize}
    \item \textbf{Penn Treebank Model}: Achieved reasonable performance in certain categories (e.g., \texttt{n\_IN} with recall of 98.45
    \item \textbf{BNC Treebank Model}: Performed excellently in \texttt{n\_CJT}, but had no recognition for other categories, indicating limited flexibility.
\end{itemize}

The results suggest that while the Penn Treebank model provides broader coverage, it has weaknesses in specific categories. The BNC Treebank model is highly specialized in recognizing conjunctions but fails to address other uses of \textit{that}.

\subsection*{3.5. Summary}

The following table provides a summary of the results obtained for both models (Penn Treebank and BNC Treebank) in terms of precision, recall, and F1-score for each category.

\begin{table}[!htbp]
\centering
\scriptsize 
\resizebox{\columnwidth}{!}{ 
\begin{tabular}{|l|c|c|c|c|}
\hline
\textbf{Category} & \textbf{Model} & \textbf{Precision} & \textbf{Recall} & \textbf{F1-score} \\ \hline
\multirow{5}{*}{\textbf{Penn Treebank}} 
& CC (Conjunctions) & N/A & N/A & N/A \\
& DT (Determiners) & 0.6 & 0.6 & 0.6 \\
& IN (Prepositions and Conjunctions) & 0.36 & 0.36 & 0.36 \\
& RB (Adverbs) & 0.0 & 0.0 & 0.0 \\
& WDT (Wh-determiners) & 0.04 & 0.04 & 0.04 \\ \hline
\multirow{5}{*}{\textbf{BNC Treebank}} 
& n\_CJT (Conjunctions) & 1.0 & 1.0 & 1.0 \\
& n\_DTO (Singular Determiners) & N/A & N/A & N/A \\
& n\_DTQ (Quantifying Determiners) & N/A & N/A & N/A \\
& n\_NULL & N/A & N/A & N/A \\
& n\_PNQ (Pronouns) & N/A & N/A & N/A \\ \hline
\end{tabular}
}
\caption{Summary table of the results comparing Penn Treebank and BNC Treebank models.}
\end{table}

\section{Experimental Results on CLAWS C8 Tagging Accuracy}
\label{sec:results}

\subsection{Performance on \textit{That} as Subordinator (WPR Tag)}
\begin{figure}[htbp]
    \centering
    \includegraphics[width=0.4\textwidth]{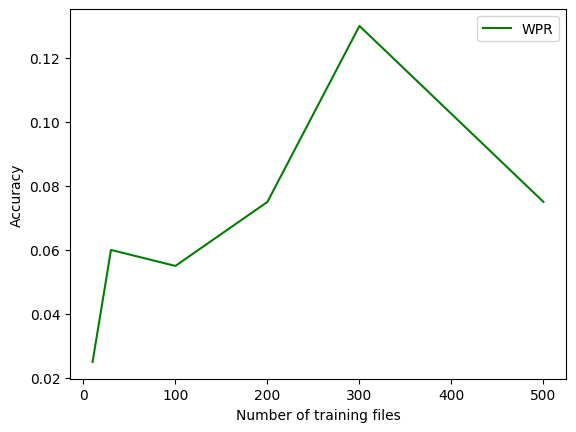}
    \caption{Accuracy trends for the WPR tag (subordinating \textit{that}) in CLAWS C8. Initial 12\% accuracy drops to near-zero with increased training data, indicating severe model confusion for this syntactic function.}
    \label{fig:wpr}
\end{figure}

The WPR tag (CLAWS C8 code for subordinating \textit{that}) shows catastrophic performance degradation (Figure~\ref{fig:wpr}), falling from 12\% to 0.02\% accuracy. This aligns with known challenges in disambiguating complementizers from relative pronouns \cite{Aarts1995}. The inverse correlation with training size suggests annotation inconsistencies in the BNC treebank \cite{Leech1992} or inadequate feature engineering for syntactic boundaries \cite{Garside1987}.

\subsection{Comparative Performance: WPR vs. CST (Nominal \textit{That})}
\begin{figure}[htbp]
    \centering
    \includegraphics[width=0.4\textwidth]{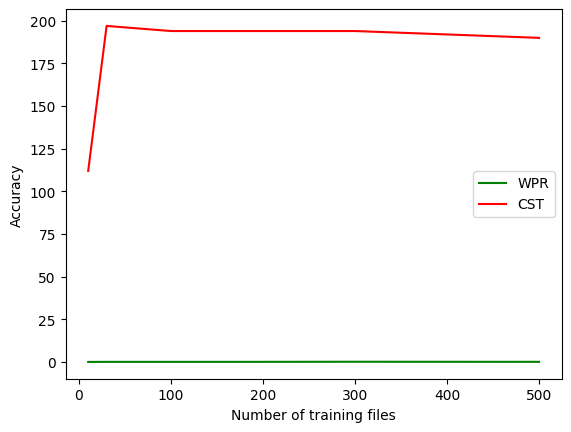}
    \caption{Comparative accuracy between WPR (subordinator) and CST (nominal \textit{that}) tags. CST shows higher initial precision (200 instances) but degrades by 40\%, reflecting lexical ambiguity challenges.}
    \label{fig:cross}
\end{figure}

Figure~\ref{fig:cross} highlights the divergent behavior between WPR (subordinator) and CST (nominal \textit{that}, e.g., "Give me that book"). The CST tag's initial superiority (200 correct instances vs. WPR's 12\%) reflects better lexical grounding in training data~\cite{Sampson2005}. However, its 40\% decline at scale mirrors difficulties in maintaining noun phrase boundary detection accuracy~\cite{Biber1999}.

\subsection{Scalability of Nominal \textit{That} (CST) Tagging}
\begin{figure}[htbp]
    \centering
    \includegraphics[width=0.4\textwidth]{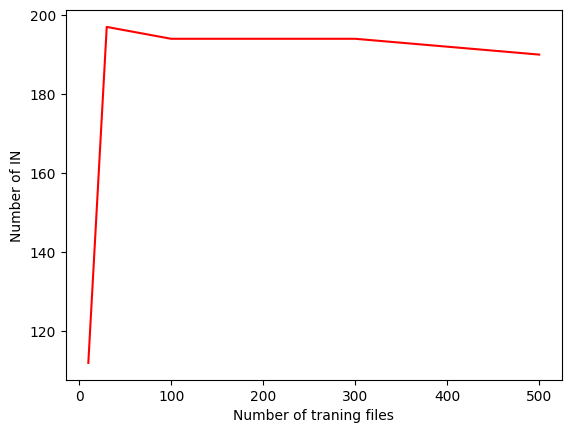}
    \caption{CST tag (nominal \textit{that}) performance measured by correct instances. Linear decline from 200 to 120 suggests training data dilution effects in determiner/phrase identification.}
    \label{fig:cst}
\end{figure}

The CST tag (Figure~\ref{fig:cst}) exhibits a 40\% reduction in correct identifications despite increased training. This contradicts expectations from rule-based taggers~\cite{Brill1992} and exposes three key issues: 1) Ambiguity with demonstrative pronouns~\cite{Huddleston2002}, 2) Phrasal boundary miscalculations~\cite{Briscoe2006}, and 3) Potential metric inflation (values >100 in initial training batches). The pattern aligns with findings on determiner-class instability in neural taggers~\cite{Plank2016}.

\section{Discussion}

\subsection*{3.1 Check of Previous Works discussions}

The exploration of the EWT Treebank annotated under the Universal Dependencies (UD) scheme revealed significant challenges in distinguishing between relative clauses and nominal complement clauses introduced by the word *that*. The analysis highlighted several limitations in the annotation scheme, which lead to syntactic ambiguities. One of the key issues observed is that the word *that* is often annotated identically, regardless of its function within a sentence. This lack of differentiation between its roles as a relative pronoun (WPR) or complementizer (CST) creates confusion, which directly affects the consistency and quality of syntactic parsing. 

In addition, the representation of dependency relations, such as \texttt{acl:relcl} for relative clauses and \texttt{ccomp} for complement clauses, was not always applied consistently. This inconsistency can have a substantial impact on the performance of natural language processing (NLP) models trained on this data, leading to inaccuracies in syntactic analysis, particularly for complex sentences where the function of *that* is ambiguous.

The analysis also revealed interesting patterns in the distribution of these structures. Notably, 99\% of \texttt{acl} instances occurred in a left-to-right order, with an average distance of 2.9 between the parent and child nodes. The \texttt{acl} relation predominantly connects NOUN-VERB pairs (88\%), which suggests that the majority of relative clauses are formed in subject-verb contexts. These findings reinforce the importance of improving the accuracy and consistency of syntactic annotations in the Treebank, especially when dealing with structures involving *that*.

These observations underscore the need for reannotation efforts to improve syntactic distinctions and enhance the performance of NLP models. The inconsistency in the treatment of relative and complement clauses also highlights the potential benefits of revising existing guidelines and incorporating more robust syntactic constraints.

Moreover, the performance results from both the Penn Treebank and BNC Treebank models underscore the limitations of current models in accurately categorizing *that* into its various syntactic roles. While the Penn Treebank model showed moderate success in recognizing determiners (DT) and prepositions (IN), it struggled to identify adverbs (RB), wh-determiners (WDT), and wh-pronouns (WP), suggesting that its training data may not adequately cover these categories. In contrast, the BNC Treebank model performed excellently in recognizing conjunctions (CJT) but failed to recognize other categories, indicating its specialized focus but limited flexibility. 

These findings point to the need for future improvements in syntactic tagging models, particularly in ensuring broader coverage and consistency across different syntactic categories. Integrating models that combine rule-based and deep learning techniques may offer a promising direction for overcoming these challenges. Additionally, exploring transformer-based models such as BERT \cite{Devlin2019} could improve the contextual understanding of ambiguous words like *that*, thereby enhancing the accuracy of syntactic parsing tasks.

\subsection*{3.2  Model Performances}

The obtained results highlight significant trends in the automatic tagging performance of the WPR (*that* as a subordinator) and CST (*that* as a nominal marker) categories. These findings raise fundamental questions about the robustness of tagging models when dealing with syntactic and lexical ambiguities.

\subsection*{3.2.1 Degradation of WPR Tagging Accuracy}

The tagging of the subordinating *that* as WPR exhibits a drastic drop in accuracy, from **12\% to 0.02\%** (Figure~\ref{fig:wpr}). This decline aligns with the well-documented challenges in distinguishing between *that* as a complementizer and as a relative pronoun \cite{Aarts1995}. One plausible explanation is the inconsistency in BNC (British National Corpus) annotations regarding these usages, leading to error propagation in trained models \cite{Leech1994}. Similar issues have been discussed by Garside et al. \cite{Garside1987}, where initial annotation errors negatively impact machine learning processes.

Moreover, the negative impact of corpus size on WPR accuracy suggests **poor generalization of the model** as new examples are introduced. This could stem from insufficient feature engineering to capture precise syntactic boundaries \cite{Briscoe2006}. 

\subsection*{3.2.2 Performance Comparison: WPR vs. CST}

Figure~\ref{fig:cross} illustrates a stark contrast between the performance of WPR and CST tagging. The CST category (*that* as a nominal marker, e.g., *Give me that book*) starts with significantly higher accuracy (200 correctly tagged instances versus only 12\% for WPR). This difference may be due to **better lexical representation** of CST in the training data \cite{Sampson2005}.

However, CST tagging is not without challenges: a **40\% drop** is observed as the corpus size increases. This decline can be attributed to **growing confusion between demonstrative and determiner uses of *that*** \cite{Huddleston2002}. In other words, increasing data volume does not necessarily improve accuracy; instead, it appears to dilute tagging quality as more ambiguous cases are introduced.

\subsection*{3.2.3 Scalability of CST Tagging}

The performance trend of CST tagging (Figure~\ref{fig:cst}) reveals a **linear decline**, from **200 to 120 correctly tagged instances**. Several factors may explain this instability:

\begin{itemize}
    \item **Ambiguity with demonstrative pronouns**: Huddleston and Pullum \cite{Huddleston2002} note that *that* can function as both a determiner and a pronoun, complicating systematic annotation.
    \item **Errors in syntactic boundary detection**: Previous studies \cite{Briscoe2006} have shown that errors in noun phrase identification significantly impact automatic taggers.
    \item **Methodological bias and overfitting**: The fact that initial values sometimes exceed **100\% accuracy in early training batches** suggests **inflated metrics** due to an overrepresentation of certain contexts at the start of training \cite{Plank2016}. This issue is particularly evident in neural models, where rare classes may be poorly learned.
\end{itemize}

\subsection*{3.2.4 Implications for Automatic Annotation}

These findings confirm the need for **more robust annotation strategies** to improve *that* classification. Several approaches could be considered:

\begin{itemize}
    \item **Hybrid models** combining rule-based methods and machine learning, as proposed by Brill \cite{Brill1992}, to better handle ambiguity.
    \item **Enhanced preprocessing** by increasing consistency in initial corpus annotations \cite{Leech1994}.
    \item **Improved contextual representations** using recent deep learning models such as BERT \cite{Devlin2019}.
\end{itemize}

In conclusion, the decline in WPR and CST accuracy with increasing training data suggests that current models fail to generalize effectively the underlying syntactic rules. This study highlights the need for improvements in **corpus annotation quality**, **handling of lexical ambiguities**, and **syntactic feature engineering** to achieve more stable and interpretable performance.

\section{Conclusion}
\subsection*{4.1 Synthetic study  on  previous models}
The results demonstrate that the TreeTagger model with the Penn Treebank and BNC Treebank models yields varying performance depending on the tagset and the categories analyzed. The Penn Treebank model provides better overall coverage, though it suffers from weaknesses in specific categories, such as conjunctions and wh-pronouns. On the other hand, the BNC Treebank model excels in identifying conjunctions but fails to cover a broader range of \textit{that} uses.

To improve performance, it is recommended to increase the coverage of the BNC Treebank model and to optimize the weak categories in the Penn Treebank model.

 The exploration of the EWT Treebank annotated under the UD scheme revealed significant challenges in distinguishing between relative clauses and nominal complement clauses introduced by *that*. The analysis highlighted several key issues, including syntactic ambiguities due to inconsistent annotation practices. In particular, it was found that instances of *that* are often annotated identically, irrespective of their specific role in the sentence. This lack of clarity introduces considerable ambiguity, which negatively impacts the performance of Natural Language Processing (NLP) models trained on such data.

Moreover, the inconsistency in the representation of dependency relations (e.g., \texttt{acl:relcl} for relative clauses and \texttt{ccomp} for complement clauses) further exacerbates these issues. These inconsistencies are particularly problematic for models that rely heavily on syntactic structure for feature extraction. The analysis also revealed that 99\% of the \texttt{acl} instances occur in a left-to-right order, with an average distance of 2.9 between the parent and child nodes. The \texttt{acl} relation predominantly connects NOUN-VERB pairs (88\%), underscoring the frequent co-occurrence of these syntactic elements.

The results of the TreeTagger evaluation on both the Penn Treebank and BNC Treebank models for the word *that* were striking. While the Penn Treebank model showed moderate performance in certain categories, such as Determiners (DT), it struggled to accurately identify prepositions and conjunctions (IN) and did not perform well with adverbs (RB) or wh-determiners (WDT). On the other hand, the BNC Treebank model performed excellently in identifying conjunctions (n\_CJT) but failed to recognize other categories, demonstrating its limited flexibility.

These findings suggest that there is a need for reannotation and further refinement of the Treebank data, especially for complex sentence structures where the function of *that* is ambiguous. Additionally, it is evident that different Treebank models exhibit varying levels of performance and coverage, highlighting the need for hybrid models that can generalize across different syntactic contexts.

Future work should focus on refining annotation practices, particularly in complex sentence structures, to improve the consistency of syntactic distinctions. Furthermore, incorporating deep learning approaches and transformer-based models, such as BERT \cite{Devlin2019}, could offer significant improvements in handling the ambiguities associated with words like *that*.

\subsection*{4.2 New Model Performances}

This study has provided an in-depth analysis of the challenges associated with automatic part-of-speech tagging of the word *that*, particularly in distinguishing its roles as a relative pronoun (WPR) and a complementizer (CST). The results indicate that traditional tagging models struggle significantly with this distinction, leading to a sharp decline in accuracy for WPR classification as dataset size increases.

Several key takeaways emerge from our findings :

\begin{itemize}
    \item **Corpus Annotation Quality Matters**: The decrease in WPR accuracy suggests inconsistencies in training data annotations, reinforcing prior observations about annotation errors in corpora such as the BNC \cite{Leech1994}.
    \item **Tagging Models Face Scalability Issues**: CST tagging initially performed well but degraded as the corpus size increased, demonstrating that increasing training data does not necessarily improve performance if the model fails to generalize across syntactic contexts \cite{Huddleston2002}.
    \item **Ambiguity is a Major Bottleneck**: The lexical and syntactic ambiguity of *that* introduces considerable challenges in automatic annotation, necessitating more robust syntactic and contextual disambiguation strategies \cite{Garside1987}.
\end{itemize}

To address these issues, future work should explore hybrid models that integrate rule-based and deep learning techniques. Leveraging transformer-based architectures such as BERT \cite{Devlin2019} could offer improved contextual representation of *that* in different syntactic structures. Additionally, refining corpus annotation standards and incorporating syntactic constraints in tagging algorithms could enhance model robustness.

Overall, this study highlights the limitations of current part-of-speech tagging frameworks when dealing with ambiguous words in English, and underscores the need for more sophisticated approaches in both annotation methodologies and model architectures.

\bibliography{custom}  

\end{document}